# Sensory Manipulation as a Countermeasure to Robot Teleoperation Delays:

# System and Evidence

Jing Du[1,*], William Vann[1], Tianyu Zhou[1], Yang Ye[1], Qi Zhu[2]

[1] ICIC Lab, Department of Civil and Coastal Engineering, University of Florida, FL 32611, USA

[2] National Institute of Standards and Technology, Boulder, CO 80305, USA

[*] eric.du@essie.ufl.edu

**Abstract**

In the realm of robotics and automation, robot teleoperation, which facilitates human-machine interaction in distant or hazardous settings, has surged in significance. A persistent issue in this domain is the delays between command issuance and action execution, causing negative repercussions on operator situational awareness, performance, and cognitive load. These delays, particularly in long-distance operations, are difficult to mitigate even with the most advanced computing advancements. Current solutions mainly revolve around machine-based adjustments to combat these delays. However, a notable lacuna remains in harnessing human perceptions for an enhanced subjective teleoperation experience. This paper introduces a novel approach of sensory manipulation for induced human adaptation in delayed teleoperation. Drawing from motor learning and rehabilitation principles, it is posited that strategic sensory manipulation, via altered sensory stimuli, can mitigate the subjective feeling of these delays. The focus is not on introducing new skills or adapting to novel conditions; rather, it leverages prior motor coordination experience in the context of delays. The objective is to reduce the need for extensive training or sophisticated automation designs. A human-centered experiment involving 41 participants was conducted to examine the effects of modified haptic cues in teleoperations with delays. These cues were generated from high-fidelity physics engines using parameters from robot-end sensors or physics engine simulations. The results underscored several benefits, notably the considerable reduction in task time and enhanced user perceptions about visual delays. Real-time haptic feedback, or the anchoring method, emerged as a significant contributor to these benefits, showcasing reduced cognitive load, bolstered self-confidence, and minimized frustration. Beyond the prevalent methods of automation design and training, this research underscores induced human adaptation as a pivotal avenue in robot teleoperation. It seeks to enhance teleoperation efficacy through rapid human adaptation, offering insights beyond just optimizing robotic systems for delay compensations.

**Introduction**

Robot teleoperation, the technique of controlling robots from a distance, has gained substantial interest within the fields of robotics and automation, facilitating human interaction with environments that are remote, hazardous, or inaccessible [1]. Through teleoperation, operators can engage in intricate tasks, surpassing the limitations imposed by the spatial separation between human and robot, and thereby, expanding the horizons of applications such as deep-sea exploration, space missions, and hazardous material handling [2-5].

A critical challenge inherent to robot teleoperation is the inevitable delays, manifesting as a latent barrier between the command issued and the corresponding action executed [6]. For example, for NASA's Space Station Remote Manipulator System (SSRMS) and the Special Purpose Dexterous Manipulator (SPDM, or Dextre), time delays can occur at different levels due to the long distances of signal transmission and limited computer processing [7]. Such teleoperation delays are caused by the physical limits that are indispensable despite improvements in computing and control efficiency. For example, in low earth orbit (LEO), the cycle time delay is at least 500ms [8], and for Earth-orbit applications it is normally between 5-10 seconds due to multiple transmission points [9]. These delays, especially in scenarios involving long-distance operations or interactions with intricate environments, become pivotal concerns, adversely impacting the operator's situational awareness, control, and overall task performance [10], leading to heightened cognitive workload and potential operational errors [11].

Mitigation strategies have been proposed to address the problems related to teleoperation delays. Representative methods include supervisory controls [6,12,13], predictive controls [9,14-16], adaptive control algorithms [17-19], and the implementation of diverse interaction modalities [20,21]. Efforts have also been made to improve manual maneuvering strategies, such as "move and wait", with excessive training [22]. However, risks of teleoperation delays still persist when the patterns of time delays can be completely unpredictable and thus designing for delays is not feasible [23], and when human operators may have little training time for emergencies [24]. Especially, while considerable advancements have been made in the development of strategies to mitigate teleoperation delays, a notable knowledge gap persists in understanding the full potentials of manipulating human operator's perceptions for a better subjective experience in delayed teleoperation. Existing research has predominantly centered on the so-called machine adaptation through predictive and supervisory controls etc., aiming to adjust system behaviors to counteract the possible delays in the control feedback loop. In other words, efforts have been made to reduce the actual delays or their impact on control [25]. In contrast, we focus on affecting the subjective experience of the human operator. We propose induced human adaptation as an alternative approach for mitigating teleoperation delays. Inspired by the motor learning and rehabilitation literature [26-28], we hypothesize that sensory manipulation [29-32], i.e., modified (time points, frequency, modality and magnitude) sensory stimulation, paired with the motor actions, helps alleviate the subjective feeling of time delays, and expedite human functional adaptation to time-delayed teleoperation, without the need for excessive trainings. Building upon neural plasticity, this approach manipulates the sensorimotor channel to expedite neural functions (motor focus) in response to changes in the environment or lesions. It is worth noting that this approach is different from training methods, as it does not require the gaining of new skillsets or adaptation to new conditions; instead, it transfers previous experience in motor coordination to the new, time-delayed, condition. As such, training needs will be reduced.

This paper aims to provide preliminary evidence about sensory manipulation as a countermeasure to the challenges posed by teleoperation delays. Specifically, we investigated whether modifying sensory stimuli simulated via high-fidelity physics engines can moderate the sense of delays and expedite the operator's adaptation to delays in manual teleoperations. A human subject experiment (N=41) was performed to examine the cognitive and behavioral implications of varied haptic cues, synchronous or asynchronous with visual cues, in time-delayed teleoperations. The haptic cues are conceptualized to be generated based on parameters procured from pressure and kinematics sensors affixed to the end effectors of the remote robot, or they can be rooted in physics engine simulations at the local workstation, providing an insightful understanding of their efficacy. Without loss of generality, this research focuses on identifying potential benefits associated with the manipulation of sensory input, a method hypothesized to alleviate the perceived delays in teleoperation tasks. The remainder of this manuscript introduces the relevant body of literature as the point of departure, the design of the sensory manipulation system, the human subject experiment, and the preliminary findings.

## Literature Review

### Robot Teleoperation Delays

Robot teleoperation, situated at the convergence of robotics, control theory, and human factors, has been pivotal in enabling human interaction with distant or hostile environments [33]. The delay in communication intrinsic to this system emerges as a critical barrier, affecting the synchronization between human commands and robotic actions [7]. These delays manifest significantly in contexts such as space exploration and underwater interventions, where substantial distances and environmental complexities necessitate extensive processing and transmission times [7,34]. The intricate nature and origin of these delays have been exhaustively studied, highlighting their profound impacts on control coherence and task performance [35].

A substantial corpus of research has substantiated that delays in teleoperation notably compromise task performance and control stability [6,36]. The resultant temporal misalignment has been shown to induce detrimental oscillations, especially in tasks necessitating high precision and prompt reactions, subsequently affecting the accuracy and prolonging completion times [37]. The exploration of mitigative strategies has been extensive, featuring developments like predictive displays and adaptive control algorithms, aiming to counterbalance the delay-induced discrepancies and instabilities [25]. However, despite these advancements, the literature indicates a prevailing need for holistic and innovative solutions to address the multifaceted challenges introduced by teleoperation delays [38].

The cognitive implications of teleoperation delays are equally significant, inducing elevated cognitive workload and impairing the learning efficacy of operators [10]. The persistent effort to reconcile anticipated and actual system states due to delays has been associated with heightened risk of operational errors and reduced operational sustainability [39]. The literature underscores the long-term impacts on operator proficiency and adaptability, emphasizing the critical role of situational awareness in effective teleoperation [40]. The disruption of temporal cohesion between perception and action, particularly in unpredictable environments, necessitates real-time situational appraisal and rapid decision-making [23].

In summary, the continuous evolution in teleoperation paradigms, marked by integrations of emerging technologies, necessitates an ongoing re-evaluation and enrichment of the literature on teleoperation delays. The developing spheres of haptic feedback, augmented reality, and advanced control algorithms are indicating novel prospects for mitigating the adverse effects of delays [41-43]. These evolving considerations underscore the imperative for groundbreaking solutions such as sensory manipulation, postulated in this paper, to ameliorate the pervasive challenges posed by teleoperation delays and to fortify the amalgamation of human cognition with robotic precision.

**Delay Mitigation**

Mitigating the inherent delays in robot teleoperation has been a focal pursuit in contemporary research, given the pivotal implications these delays impose on overall system performance and operator experience. A seminal approach in this context is the incorporation of predictive controls [9,14-16]. This methodology, extensively evaluated by researchers like Uddin and Ryu [25], emphasizes providing operators with anticipatory visual cues, allowing an interpretation of anticipated robotic actions prior to actual system responses. The approach utilizes intricate mathematical models to simulate future system states, optimizing operators' adaptability and situational awareness in delay-prone environments and minimizing the discord between anticipated and actual system states, a vital step in optimizing task performance [44].

Parallelly, the integration of supervisory controls has been explored as a robust solution to mitigate the multifaceted impacts of teleoperation delays [6,12,13]. This modality allows for an intelligent delegation of control tasks to the autonomous subsystems within robots, reducing the necessity for continuous manual input and mitigating the adverse impacts of delays on operator workload and task efficacy [45]. Furthermore, adaptive and robust control strategies have been at the forefront of mitigative research, focusing on maintaining system stability and performance optimization amidst varying operational conditions by dynamically adjusting control parameters in alignment with observed system states and delay magnitudes [18].

Additionally, the realm of sensory feedback and multimodal interaction has undergone extensive exploration, seeking to enhance the operator's perceptual awareness and response to delays. Pioneers in this domain, such as Massimino and Sheridan [46], advocate for the integration of diverse interaction modalities and the manipulation of sensory inputs, aiming to foster a more intuitive and immersive operator interaction experience. This enhanced interaction paradigm facilitates more effective operator adjustments in response to delays and, when synergized with other mitigative strategies, opens avenues for a holistic approach to delay mitigation [47]. The ongoing pursuit for mitigative strategies underscores the collective aspiration of the scholarly community to explore innovative, adaptive, and integrative solutions to counter the complexities and challenges induced by teleoperation delays. The emerging consensus emphasizes a multifaceted approach, amalgamating predictive and supervisory controls, adaptive algorithms, and enhanced sensory feedback, striving to construct a comprehensive mitigative framework that addresses the myriad facets of teleoperation delays in a cohesive and synergistic manner.

**Sensory Manipulation**

The exploration of sensory manipulation unfolds as a pivotal frontier in mitigative strategies for teleoperation delays, spotlighting the intricate intertwining of sensory perception, cognitive psychology, and robotic control [46]. This technique focuses on modifying sensory stimuli paired with the operator's motor actions to optimize perception and reaction in delayed teleoperations, functioning as a cognitive countermeasure to discrepancies between expected and perceived system responses [48].

A salient concern addressed by sensory manipulation is the induced perceptual-motor malfunction arising from time delays in motor action, characterized by the inability to effectively integrate perceptual information with the execution of voluntary behaviors [49-51]. This is profoundly consequential as human sensorimotor control is inherently reliant on multimodal sensory feedback, encompassing visual, auditory, and somatosensory (tactile and proprioceptive) cues, to decipher the consequences of the initiated action [52-54]. When perceptual ability is compromised, it results in a broken motor planning and feedback loop. Such perceptual-motor malfunctions are discernible in clinical populations with impaired perceptual functions, like Asperger disorders, Parkinson's disease, and Developmental Coordination Disorders (DCD), exhibiting particularly heightened challenges in visual, spatial, and tactile domains [55-57].

In the realm of teleoperations, the perceptible lags between motor action and feedback create analogous mismatches in motor perception, leading to comparable consequences of perceptual-motor dysfunction. To address this challenge, extensive literature in learning and rehabilitation accentuates the impact of modifying sensory stimuli from the surroundings. It has been established that providing associated visual, auditory, and haptic cues with an intended action not only influences motor performance but also modulates the efficacy of motor rehabilitation [26-28]. The robust manipulation of sensory information during motor tasks has paramount implications for enhancing motor learning across healthy individuals and clinical populations, leveraging the potential of sensory manipulation to mitigate perceptual-motor malfunctions effectively [29-32]. Furthermore, the manipulation of haptic cues has been explored extensively, focusing on their synchronous or asynchronous integration with visual cues. Studies such as [58] highlight the substantial cognitive and behavioral benefits of this modality, revealing the potential for enhanced adaptation to delays in teleoperations and refined task performance and situational awareness. Enhanced haptic feedback, whether based on real-time sensor data or physics engine simulations, emerges as a versatile and impactful application in diverse teleoperation contexts, contributing significantly to the advancement of holistic and integrative solutions for teleoperation delays.

In summary, the exploration and application of sensory manipulation echo as a promising solution in the multifaceted journey to mitigate teleoperation delays. By forging cohesive connections between diverse sensory inputs and leveraging advanced interaction modalities, sensory manipulation empowers operators with refined adaptation mechanisms and enriched perceptual awareness in the face of delays. This innovation illuminates unprecedented avenues, heralding a transformative era in robot teleoperation marked by enhanced resilience and integrative adaptability.

## Design of the Sensory Manipulation System

### System Architecture

**Fig.1** illustrates the system architecture of the proposed sensory manipulation system for providing augmented sensory cues, especially haptic feedback, for robot teleoperation with varying delays.

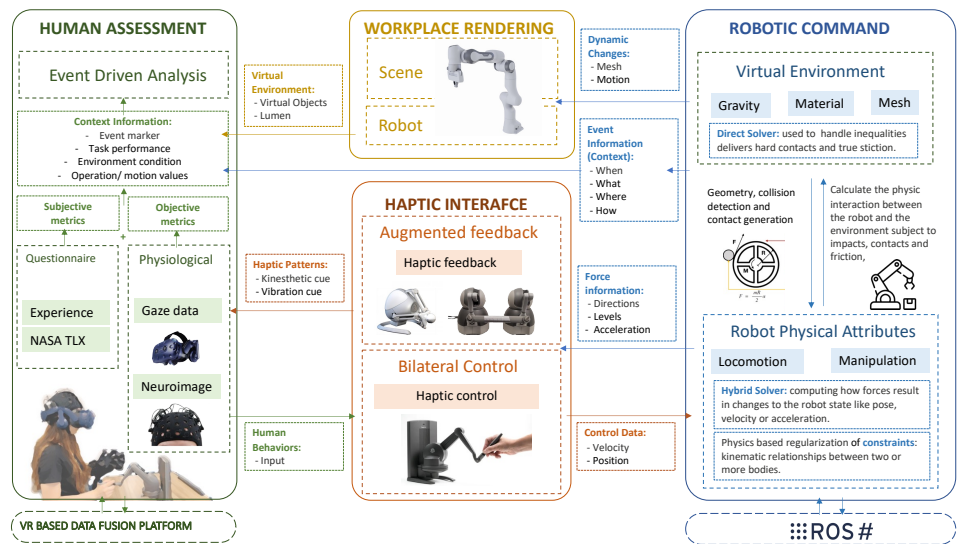

**Fig.1** Architecture of the sensory manipulation system for robot teleoperation

Specifically, the system includes the following four main units: **(1) Robot commanding unit (RC).** The RC unit connects a robot arm (either a real system or simulated model) with the Unity game engine for digital twin simulation and for haptic controls. Robot operating system (ROS) is used as the main platform for exchanging data between the ROS system and Unity. This component is also responsible for converting the control commands into the locomotion topics of the robot using inverse kinematics (IK) algorithms. **(2) Workplace rendering unit (WR).** Unity game engine is used to create a digital twin model of the remote robot and the workplace. Human operators can use a VR headset to visualize the remote workplace and the robot for coordinating the hand-picking tasks in an immersive way. **(3) Haptic interface unit (HI).** It includes haptic feedback and control systems. A total of seven types of physical interactions, including *weight*, *texture*, *momentum*, *inertia*, *impact*, *balance*, and *rotation* are simulated via a physics engine, and then are played via a high-resolution haptic controller. To be noted, we also programmed the system to intentionally add levels of latencies to the visual or haptic feedback. **(4) Human assessment unit (HA).** The last component of the system includes a set of neurophysiological sensors embedded in the VR system for real-time human assessment, including eye trackers, motion trackers, and functional near-infrared spectroscopy (fNIRS) to examine the hemodynamic activities in brain regions of interest (ROIs). This unit also includes a set of subjective questionnaires, such as NASA TLX [68] and experience surveys, to include the perception data in the final collection. The following sections introduce the technical details of each unit.

### Robot Commanding Unit (RC)

Enabling seamless control data exchange between human operators and the remote robot is crucial for realizing the proposed sensory manipulation system for robot teleoperation. We utilized ROS and Unity game engine to facilitate the RC functions, with Unity specializing in real-time modeling of human motion data and providing a robust platform for developing immersive and intuitive VR working spaces for operators. Haptic device control data from the human operator is captured and processed in Unity before being streamed to ROS for real-time robot control, based on our previous works [1,69]. The Emika Panda robot is used to as the robotic model in the proposed architecture. **Fig.2** presents the ROS-Unity data synchronization system architecture.

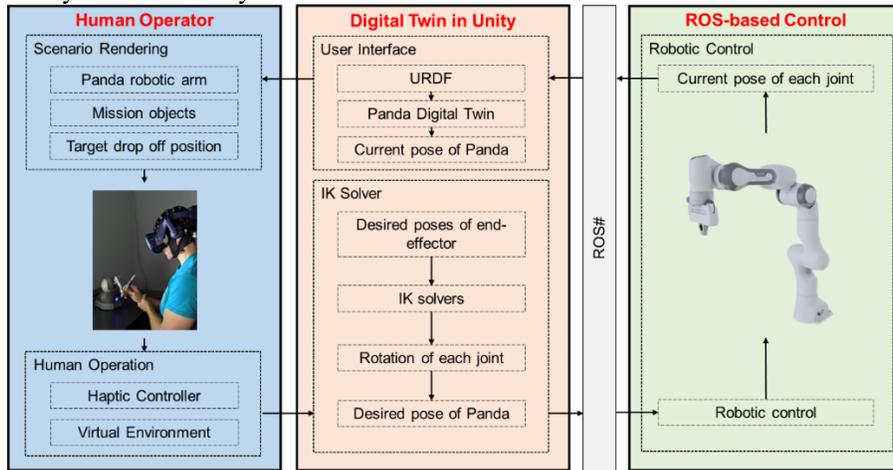

**Fig.2** ROS-Unity system

**Fig.2** depicts the three main functions of the RC unit: ROS-based control, digital twin reconstruction in Unity, and human mechanics capturing and conversion. ROS handles the input commands and publishes the robot's state to establish two-way communication. The communication between Unity and ROS is facilitated by ROS# and ROSbridge, enabling data transmission via public networks and ensuring interaction between ROS and Unity [70,71]. The ROS server converts and publishes robotic dynamics data, further used to control the robot. On Unity's end, ROS# allows the construction of compatible nodes and establishes a WebSocket for data exchange with .NET applications [71], ensuring seamless data sharing between Unity and ROS. A digital twin of the robot arm, built from a Unified Robot Description Format (URDF) file [72], receives data and behaves compatibly with the real robot, with Unity subscribing to real-time location and orientation data of each robot joint. Human operators operate a haptic controller (to be discussed later) to manipulate the robot's end effector in the virtual models in a virtual environment. The control data is transferred to Unity for processing, and an IK solver recovers the desired robot state, transmitted to ROS for real robot control [73]. Unity subscribes to the converged real robot state to update the digital twin model, closing the loop by visualizing it to users. Our system utilizes a TCP/IP protocol of ROS, i.e., TCPROS [74], for data transfer between different terminals and processing multi-processes of robot controls and 3D scene data simultaneously. As illustrated in **Fig.3**, TCPROS is a transport layer of ROS Messages and Services using standard TCP/IP sockets for transferring message data [74], including header information with message data type and routing information. Using TCPROS, our system connects processes of programs or nodes that execute different functions, such as robot controls.

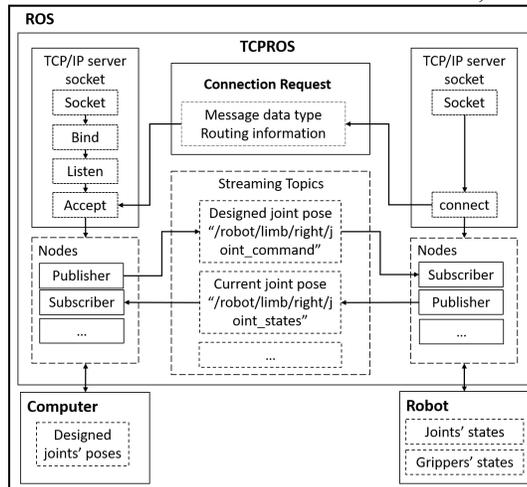

**Fig.3** Data Synchronization in ROS

## Workplace Rendering Unit (WR)

The Workplace Rendering Unit (WR) serves as a pivotal component in our teleoperation system, utilizing the Unity game engine to precisely simulate the remote robot and its workplace, thereby rendering an immersive and interactive model. This simulation model is achieved through integration of varied pipelines in Unity, notably the FBX [75], known for its capability to produce photorealistic renderings, thereby allowing for enhanced environmental immersion and user interaction. This integrative approach ensures detailed visual representation of both the robotic elements and their surrounding workspace, reflecting real-world nuances with high fidelity. In addition, the rendering precision achieved in Unity is complemented by the incorporation of robust physics engines, such as PhysX [76], critical for simulating physics-accurate environments. This engine facilitates the incorporation of intricate environmental variables and dynamic elements, allowing the simulation to reflect behaviors consistent with physical behaviors and to emulate varying gravitational conditions with accuracy. The laws governing these physical behaviors are solidly grounded in Newton's Laws of Motion, which, as established by Bregu, et al. [77], articulate the fundamental relationships between an object's motion and the forces applied upon it. Such physics-based interactions are crucial for delivering a realistic user experience, especially for tasks necessitating precise interaction with the environment, as it provides the human operators with a reliable and coherent representation of the remote workplace dynamics. Another key requirement for the WR unit is the seamless rendering of the robotic system. Herein, the robot's specific attributes and the URDF are used for crafting a virtual counterpart that mirrors the real robot's states, ensuring accuracy and synchronization between the virtual and physical entities [72]. This sophisticated integration of Unity and PhysX not only focuses on rendering but is also developed to simulate physics-accurate environments. This approach guarantees that every interaction within the simulated environment adheres strictly to established principles of classical mechanics, hence rendering the interactive experience exceptionally realistic and reliable. **Fig.4** shows the screenshot of the simulated environment.

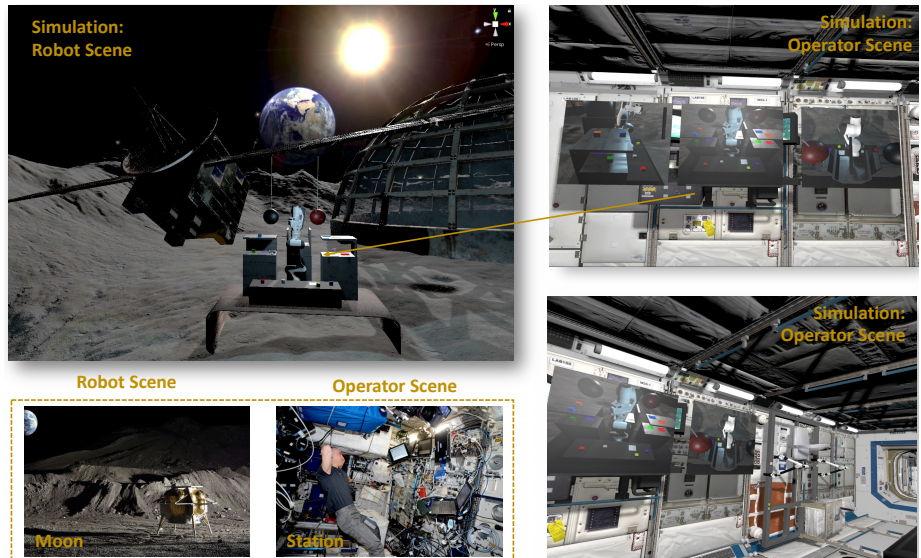

**Fig.4** The developed VR environment for this research

To enhance the user experience and task coordination, human operators are enabled to visualize the remote workplace and the robot using a VR headset, such as HTC VIVE. This feature transforms the teleoperation experience by immersing the operators within the simulated environment, allowing them to coordinate hand-picking tasks intuitively. By placing operators inside a detailed, physics-accurate, and photorealistic rendering of the remote environment, the system offers unparalleled insight and control over the remote tasks, enabling operators to execute tasks with enhanced precision and situational awareness.

## Haptic Interface Unit (HI)

To realistically simulate haptic feedback in the system, we designed six physical force modes: *weight*, *texture*, *momentum*, *inertia*, *impact*, and *balance*, each modeled in Unity using various mathematical formulations adapted to our teleoperation system and haptic device. For a detailed technical description, please refer to our previous publication [1]. These modes of physical feedback are then played via a high-resolution haptic controller, in our study, TouchX [78]. TouchX is as an epitome of modern haptic technology, equipped with high-precision force feedback capabilities that provide an immersive tactile experience. With a spatial resolution of less than 0.1mm and a force feedback range up to 5N, TouchX can reproduce minute details, allowing users to genuinely "feel" virtual physical interactions [78]. The controller is further enhanced with multi-degree-of-freedom movements, ensuring the user gets a holistic haptic experience [78]. The integration of TouchX with our system, combined with the aforementioned physical force modes, ensures that the user gets the most accurate and realistic haptic feedback possible. These modes create an embodied experience, with the force feedback being generated based on adapted Newton's laws of motion and specific equations tailored to suit different force types, thereby ensuring that the human operator receives precise and realistic tactile feedback corresponding to the various physical forces being simulated. These adapted equations, implemented directly through the ROS and Unity, facilitate smooth communication between our high-level control software and the physical robot hardware, allowing the operator to control the robot intuitively while simultaneously receiving accurate force feedback. **Fig.5** shows the representative scenarios of these simulated physical feedback modes.

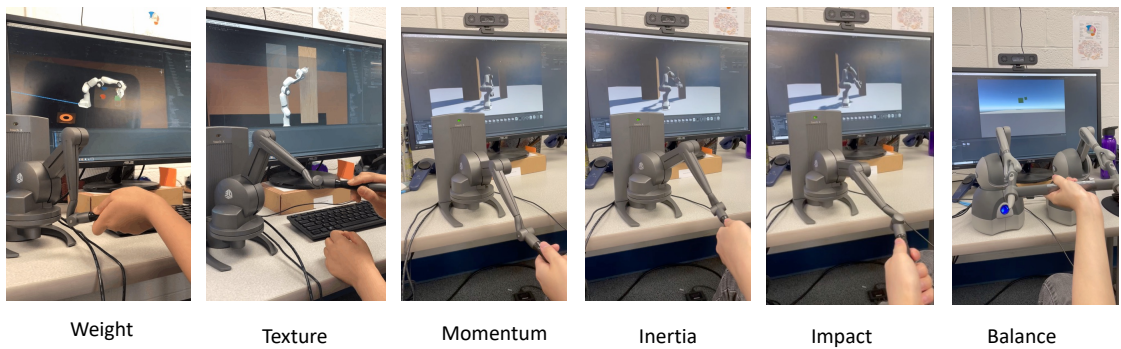

Weight    Texture    Momentum    Inertia    Impact    Balance

**Fig.5** Augmented physical feedback modes via our sensory manipulation system

Our system uses the Unity game engine [79], known for its robust physics engine, to manage physics simulations, create 3D virtual environments, and control interaction logic for robot teleoperation. The integration of ROS provides essential services like hardware abstraction, low-level device control, and the implementation of commonly used functionality. This integration ensures that the various force modes are rendered in real-time, with each mode having its unique property and equation, allowing the operators to experience high-resolution interactions with the virtual environment. The motion of the haptic device is then translated into corresponding movements of the robot's end-effector through IK [73], which is essential in enabling the human operator to control the robot's pose and grabber switch state, responding differently based on the visual and haptic feedback received due to the varying properties of objects made of different materials.

Moreover, to enhance the interactive capabilities within virtual environments, we developed a two-way communication device by combining the TouchX haptic device with an Arduino pressure sensor. This combination enables the simultaneous reception of force feedback and transmission of pose and pressure data. The TouchX haptic device is pivotal in reproducing the sensed forces and controlling the joint poses of the remote robot, while the Arduino pressure sensor is crucial for sensing the human operator's grasping force to control the gripper of the robot. Several modifications and additions, including the use of moldable silicone, hard plastic boards, and mounting tape, were incorporated to ensure the stability and reliability of the system, considering the uniqueness in each individual's finger shape and force application technique.

The HI unit also enables bilateral control, where the human operator can move the haptic controller handler to control the end effector of the remote robot in a natural way. To be noted, our system provides augmented haptic feelings (such as grabbing a weight in hands or hitting a heavy object) in addition to regular tactile stimulations. In this system, we also programmed the system to intentionally add four levels of latencies – 250ms, 500ms, 750ms, 1000ms - to the visual or haptic feedback. As a result, we can test how different levels of latencies affected the teleoperation performance with our sensory manipulation system.

## Human Assessment Unit (HA)

The HA unit leverages a sophisticated API-based system meticulously designed to autonomously harvest motion data from body-carried HTC VIVE motion trackers and HTC VIVE Eye Pro eye trackers. This system is an augmentation of our well-validated VR systems, which integrates both eye-tracking and motion tracking functionalities to capture high-precision and high-resolution gaze movement data, employing the Tobii Pro eye tracker synergized with HTC VIVE Head Mounted Display (HMD).

The Tobii Pro VR integration, developed by Tobii, employs the avant-garde Pupil Centre Corneal Reflection (PCCR) remote eye-tracking technique [80]. This technique utilizes near-infrared illuminators to create reflection patterns on the eye's cornea and pupil, and cameras within the eye tracker capture high-resolution images of these patterns. These images, when processed through advanced image-processing algorithms and integrated with a physiological 3D model of the eye, enable the accurate estimation of the eye's position within the virtual environment and the determination of the user's pupil size [81]. This sophisticated integration offers an unparalleled accuracy of 0.5° and can output gaze data at a maximum frequency of 120 Hz. To implement eye-tracking and playback functionalities within the virtual environment, we crafted several C# scripts, interlacing them with the Tobii Pro Software Development Kit (SDK) and the application programming interface (API) in Unity [82]. The environment is thus capable of collecting gaze movement data and pupil diameter data at a frequency of 90 Hz. These gaze and pupil tracking metrics act as auxiliary evidence in the comprehensive stress assessment procedure [81].

Additionally, the system is adept at assessing body and hand movement data to inspect task performance dynamically. This incorporates technical methods such as inertial measurement units (IMUs) for precise body movement analysis, and spatial analysis algorithms to assess and evaluate the complexity and efficiency of hand movements. The seamless fusion of these technologies allows for exhaustive data collection, including variations in movement speed, trajectory, and accuracy, enabling nuanced evaluations of task execution. Post each VR experimental trial, the system autonomously compiles and generates a CSV file containing all the raw data, allowing for in-depth post-analysis [81].

## Human Subject Experiment

The study was approved by the Institutional Review Board (IRB) of the University of Florida, Gainesville, FL, USA (No. IRB202100257). Written informed consents were obtained from all participants in full accordance with the ethical principles of the relevant IRB guidelines and regulations. All methods were carried out in accordance with relevant guidelines and regulations. The following inclusion criteria were applied: (1) age ≥ 18 years; (2) no known physical or mental disabilities; (3) no known musculoskeletal disorders.

## Experiment Task

The core task in our human-subject experiment was an object manipulation task. Participants were presented with four distinctively colored cubes: purple, grey, blue, and green. Each cube was associated with a corresponding target, and participants were required to relocate these cubes to their respective targets. The movement of the cubes was not arbitrary but followed a pre-scripted sequence: grey, green, blue, and then purple. This sequence was intentionally designed to represent escalating levels of difficulty. The distances between each cube and its target varied, with obstacles strategically placed along the movement trajectories. These obstacles, differing in their sizes and positions, added layers of complexity to the task, representing varying movement challenges that participants had to navigate. **Fig.6** illustrates the layout of the experiment task.

During the experiment, participants took control of the robot's gripper to grasp the cubes. Once secured, they had to maneuver the cubes past the aforementioned obstacles and accurately place them onto their corresponding target plates. The precision with which the cubes were positioned on the targets was crucial, as this was a key metric in evaluating the participants' manipulation performance. Compounding the challenge, the task was not always straightforward. Participants had to perform their manipulations under different conditions, marked by varying levels of visual and haptic delays. These delays, described in subsequent sections, were introduced to understand the participants' adaptability and proficiency under diverse sensory manipulation conditions.

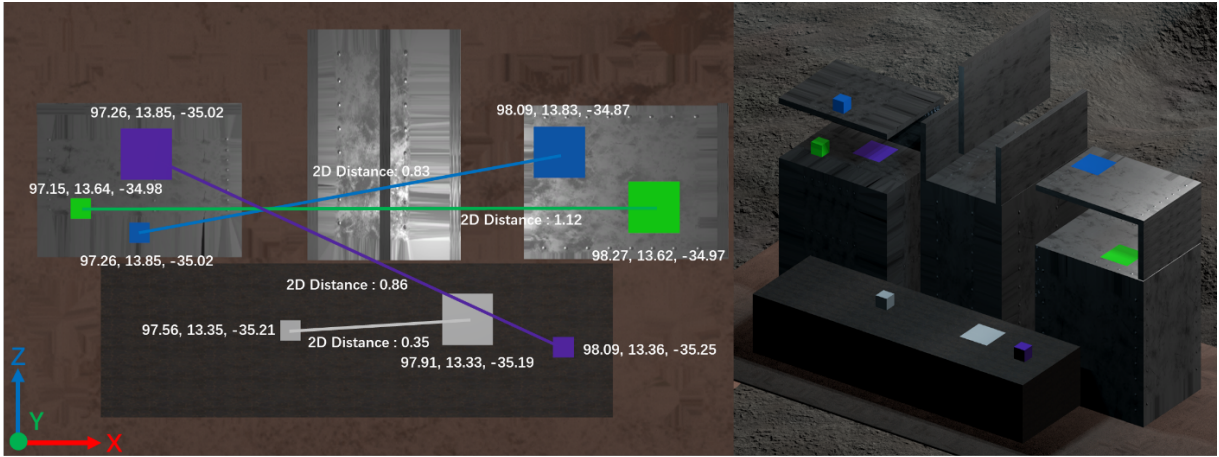

**Fig.6** The layout of the object manipulation task in human-subject experiments

**Experiment Design**

The haptic simulation reproduces the contact dynamics of the remote robotic system (e.g., resistance, torque, and nominal weight etc.) for operator via haptic devices. Based on how the haptic simulation was modified (in terms of timing and modes), four conditions were tested, as shown in **Fig.7**.

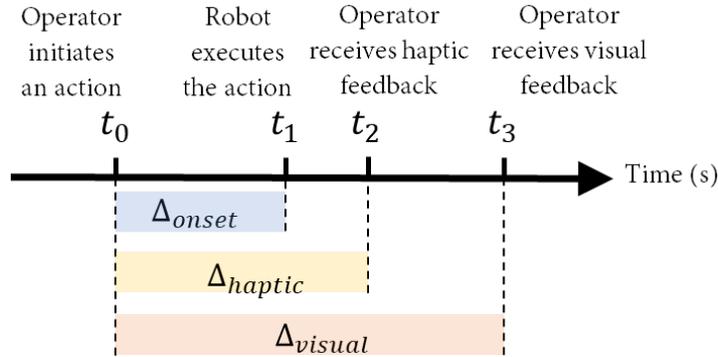

**Fig.7** Design of time delays. In an ideal situation, $\Delta_{onset} = \Delta_{haptic} = \Delta_{visual} \cong 0$. The most likely situation is $0 < \Delta_{onset} < \Delta_{haptic} < \Delta_{visual}$ (visual data can be bigger).

The total delay is decomposed into *onset delay* ($\Delta_{onset}$), *haptic feedback delay* ($\Delta_{haptic}$), and *visual feedback delay* ($\Delta_{visual}$). $\Delta_{onset}$ is the time difference between an operator initiates an action and when the robot executes the action. It may be caused by the command transmission lag from operator to the remote robot, and/or time required for running inverse kinematics to calculate variable robot joint parameters [83]. $\Delta_{haptic}$ and $\Delta_{visual}$ are feedback delays, representing times needed for the operator to receive haptic and visual feedback respectively. Based on the analytical framework, the four sensory manipulation conditions are:

**Condition 1: Control condition = $\Delta_{haptic} = \Delta_{visual}$ (*haptic feedback=0; visual feedback=0*).** In this condition, haptic and visual feedback both happen immediately after an action initiated by the human operator, i.e., in real time.

**Condition 2: Anchoring: $0 \cong \Delta_{haptic} < \Delta_{visual}$(*haptic feedback=0; visual feedback=250ms, 500ms, 750ms, and 1000ms*).** This condition generates haptic stimulation immediately after the operator initiates an action. However, it should be noted that in most cases $0 < \Delta_{onset} < \Delta_{haptic}$. As a result, the parameters for haptic feedback will be based on simulation. Specifically, the haptics can be constant vibrations to indicate that the motion has begun, or can be simulated force feedback (e.g., inertia, resistance, simulated virtual weight, and major contact events) based on the physics engine simulation at the local workstation. The operator receives haptic feedback immediately after initiating the action, but still needs to wait for delayed visual cues (e.g., camera view) due to transmission.

**Condition 3: Synchronous: $0 < \Delta_{haptic} = \Delta_{visual}$ (*haptic feedback=visual feedback=250ms, 500ms, 750ms, and 1000ms*).** In this condition, the haptic feedback is intentionally delayed to match with the delayed visual feedback. If $\Delta_{haptic} < \Delta_{visual}$, it can be achieved by adding a time buffer to $\Delta_{haptic}$. The haptic feedback is based on parameters obtained from the sensors attached to the end effectors of the remote robot and thus is more precise. The rationale is to ensure *multisensory congruency*, i.e., a coherent representation of sensory modalities to enable meaningful perceptual experiences [84].

**Condition 4: Asynchronous: $0 < \Delta_{haptic} < \Delta_{visual}$ (haptic feedback=250ms; visual feedback=250ms, 500ms, 750ms, and 1000ms).** This condition reflects the most likely and the most challenging time delay scenarios. There are perceivable delays between the action initiation and the haptic feedback, and between the haptic and visual feedback.

The experiment was designed as a within-participant experiment, i.e., each participating subject experienced four conditions. To avoid learning effects, the sequence order was shuffled for each subject. The performance data (time and accuracy), motion data (moving trajectory), eye tracking data (gaze focus and pupillary size), and neurofunctional data (measured by Functional Near-Infrared Spectroscopy or fNIRS) were collected. Participating subjects were also requested to report their perceived delays, to compare with the actual delays. Before experiment, each participant was required to fill out a form of demographic survey, and the consent form approved by UF's IRB office. Then they would take a training session, to familiarize with the use of VR, for 10 minutes. Afterwards, participants were required to take a break of 5 minutes by sitting quietly with all sensors on. This break session was for collecting baseline data (e.g., pupillary diameter and fNIRS baseline), and to remove possible impacts of the training session. After each experiment trail, participants were promoted to fill out questionnaires related to NASA TLX and trust.

## Results

### Participants

We successfully recruited 41 healthy subjects (college students) to participate in this experiment. Among all participants, there were 26 males (63.41%) and 15 females (36.59%). The average age was 25.12 (σ=9.34). A total of 18 were with engineering background, majored in civil engineering or mechanical engineering, while others were self-identified as non-engineering students. In addition, 12 participants (29.27%) self-identified themselves with previous experience with VR, while others claimed lack of experience with VR. To be noted, our post-experiment analysis did not find any difference among these demographic or experience groups. **Fig.8** shows a participant in the experiment.

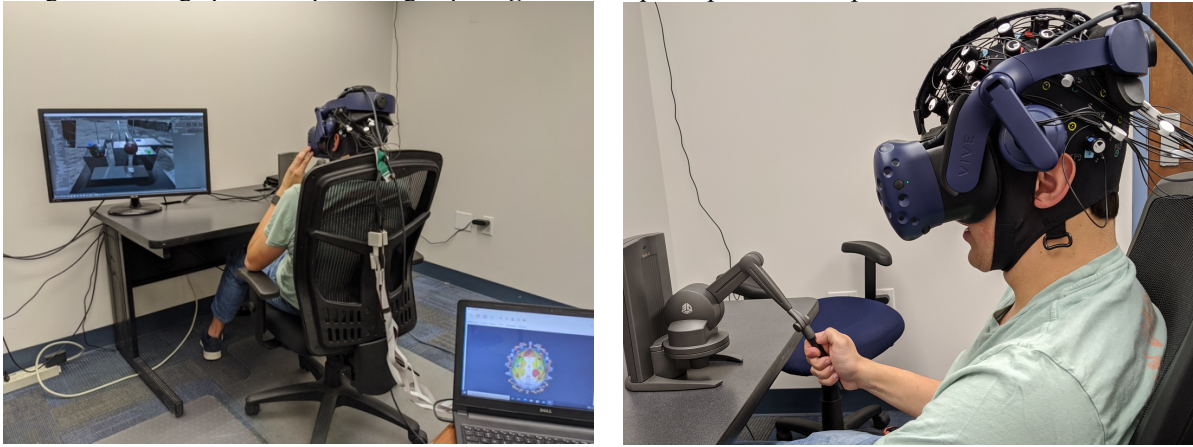

**Fig.8** A participant performing the task.

### Performance

We first analyzed the performance of the object manipulation, including time on task measured in seconds, and placement accuracy measure in cm. For placement accuracy, we can measure it as the Euclidean distance between the actual placement of the cube and the center of the target location. The smaller this distance, the more accurate the placement, as shown in Eq (1).

$$Placement\ Accuracy\ (PA) = \sqrt{(x_t - x_0)^2 + (y_t - y_0)^2} \qquad \ldots Eq(1)$$

Where: ($x_t$, $y_t$) are the coordinates of the actual placement of the cube, and ($x_0$, $y_0$) are the coordinates of the center of the target location. Time on Task is the difference between the end time and the start time of the task, as shown in Eq (2).

$$Time\ on\ task\ (ToT) = t_{end} - t_{start} \qquad \ldots Eq(2)$$

Where: $t_{start}$ is the time at which the participant grabs the cube, and $t_{end}$ is the time at which the participant drops the cube. **Fig.9a** shows the result of placement accuracy, while **Fig.9b** shows the time on task among four conditions. Without further notes, all analyses are based on the aggregated data of four cubes.

As shown in **Table 1**, the results indicate significant differences of placement accuracy between the control condition and asynchronous condition (p=0.007), between the control condition and the synchronous condition (p=0.004), between the anchoring condition and the asynchronous condition (p=0.043), and between the anchoring condition and the synchronous condition (p=0.032). Other than the control condition, we found that the anchoring condition, i.e., providing haptic cues coupled with a motor action, significantly improved the hand-picking task in terms of placement accuracy, independent of the visual delays. The anchoring condition uses simulated haptic feedback to augment a person's motor action despite visual delay levels. The benefits could be because participants could rely more on haptic feedback when it was available to coordinate the teleoperation actions. The benefit of providing a real-time haptic stimulation boosted the performance to a level similar to the control condition (i.e., no delay) (p=0.168). As for time on task, the results also indicate significant differences between the control condition and asynchronous condition (p<0.001), between the control condition and the synchronous condition (p<0.001), between the anchoring condition and the asynchronous condition (p=0.018), and between the anchoring condition and the synchronous condition (p=0.049). While the anchoring condition didn't perform as well as the control condition (p=0.009), it still outperformed both the asynchronous and synchronous conditions in terms of time on task.

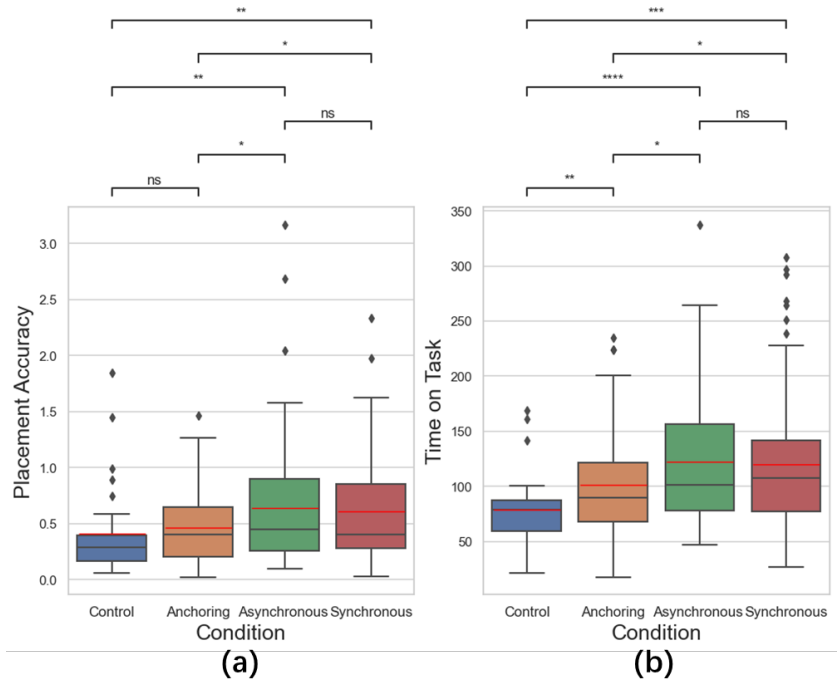

**Fig.9** Teleoperation (a) placement accuracy and (b) time on task comparison

| Condition Comparison | Placement Accuracy | Time on Task |
| --- | --- | --- |
| Control vs Anchoring | No Difference (p=0.168) | Smaller (p=0.009) |
| Control vs Asynchronous | Smaller (p=0.007) | Smaller (p<0.001) |
| Control vs Synchronous | Smaller (p=0.004) | Smaller (p<0.001) |
| Anchoring vs Asynchronous | Smaller (p=0.043) | Smaller (p=0.018) |
| Anchoring vs Synchronous | Smaller (p=0.032) | Smaller (p=0.049) |
| Asynchronous vs Synchronous | No Difference (p=0.892) | No Difference (p=0.741) |

**Table 1** Statistical results of the performance metrics

**Perception**

Then we analyzed the perception performance. We focused on examining three time perception metrics: visual perception difference, haptic perception difference, and visuomotor gap perception difference. Visual perception difference is defined by the difference between the perceived visual delay ($Delay_{vp}$) and the actual visual delay ($Delay_{va}$) in a trial, i.e.,

$$\Delta_v = Delay_{vp} - Delay_{va} \qquad \ldots Eq(3)$$

Haptic perception difference is defined by the difference between the perceived haptic delay ($Delay_{hp}$) and the actual haptic delay ($Delay_{ha}$) in a trial, i.e.,

$$\Delta_h = Delay_{hp} - Delay_{ha} \qquad \ldots Eq(4)$$

Note there were cases when there was a gap between the visual delay and the haptic delay, which we call visuomotor gap. We are also interested in the perception of the visuomotor gaps in different conditions. Visuomotor perception difference is defined by the difference between the perceived visuomotor gap ($Gap_p$) and the actual visuomotor gap ($Gap_a$) in a trial, i.e.,

$$\Delta_{gap} = Gap_p - Gap_a \qquad \ldots Eq(4)$$

**Fig.10** and **Table 2** show the results of perception performance. The results suggest that the proposed sensory manipulation method could also reduce subjective feeling of teleoperation delays (visual delays) up to 1s. Here we focus on examining if the proposed sensory manipulation method could reduce perceived visual delay, as it is considered as the most common and the most troublesome delay in robot teleoperation. The data shows that under the anchoring condition, the overall average perceived visual delay in teleoperation was significantly lower than the synchronous condition. In addition, 18% of participants reported a perceived visual delay that was smaller than the actual one under the anchoring condition. Knowing that both the anchoring condition and synchronous condition feature fixed haptic feedback after a motor action (either in real time or after 250ms), it means that coupling real-time haptic feedback with the action during teleoperation can mitigate the subjective feeling of delays. For example, when the actual visual delay was 750ms, a subject reported 100ms as the perceived delay. **Fig. 3b** and **Fig.3c** shows the comparison about the haptic visual perception difference and visuomotor gap perception difference. As for the perceived haptic delays, the data shows a little different pattern. Subjects seemed to report a lower perceived haptic delay under the synchronous condition. This makes sense because the coupled haptic and visual feedback may help a better estimate of the haptic delay. As for the visuomotor gap perception, it shows that under the anchoring condition, a significant amount of subjected reported a delay smaller than the actual one. All these results confirmed the perceptual benefits of having haptic feedback synchronous with the action.

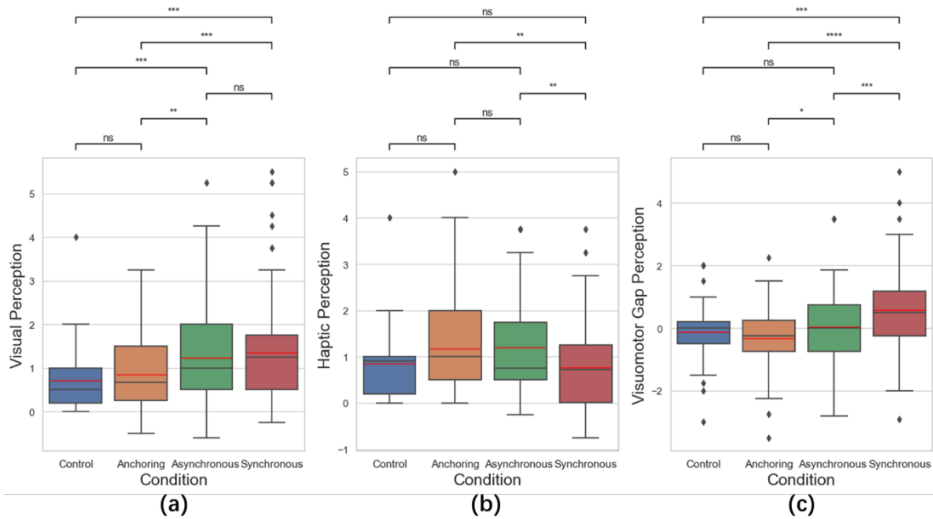

**Fig.10** Perception performance. (a) visual perception difference; (b) haptic perception difference; (c) visuomotor gap perception difference.

| Condition Comparison | Visual Perception | Haptic Perception | Visuomotor Gap Perception |
|---|---|---|---|
| Control vs Anchoring | No Difference (p=0.448) | No Difference (p=0.091) | No Difference (p=0.237) |
| Control vs Asynchronous | Smaller (p<0.001) | No Difference (p=0.090) | No Difference (p=0.534) |
| Control vs Synchronous | Smaller (p<0.001) | No Difference (p=0.052) | Smaller (p<0.001) |
| Anchoring vs Asynchronous | Smaller (p=0.003) | No Difference (p=0.098) | Smaller (p=0.024) |
| Anchoring vs Synchronous | Smaller (p<0.001) | Larger (p=0.003) | Smaller (p<0.001) |
| Asynchronous vs Synchronous | No Difference (p=0.506) | Larger (p=0.001) | Smaller (p<0.001) |

**Table 2** Statistical results of the perception metrics

## Cognitive Load

We are also interested in examining the cognitive dynamics during the course of the experiment. In addition to the NASA TLX at the end of the experiment trials, we developed a novel approach to evaluate participants' real-time cognitive load based on their pupillary diameter data collected by eye trackers, as illustrated in **Fig. 11**. This approach allowed us to capture participants' cognitive load status during the key phases of the task, such as picking up and dropping off cubes.

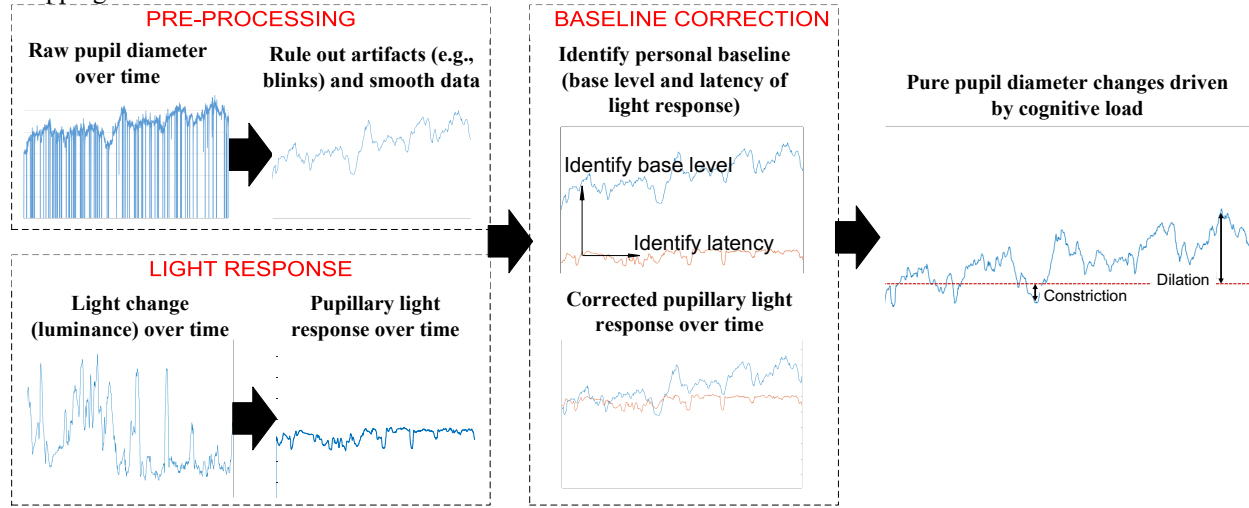

**Fig. 11** Workflow of pupillary diameter analysis

We started by correcting pupillary blink responses in the raw pupil diameter data, identifying and rectifying blink patterns within 400 - 600 ms using a linear interpolation method. Recognizing that increased motor task complexity can enlarge pupil diameter while heightened motor task precision reduces it during response planning and execution, our experiment was structured to neutralize these effects. During the pre-experiment phase, participants were promoted to take a break session for baseline measurement. In the experiment phase, consistent task complexity was ensured for all groups through the same manipulation task, while a novel "invisible collider-box method" guaranteed uniform precision measurement for the placement phase. This method utilized an invisible boundary in a virtual environment, compelling participants to maintain consistent precision levels. For data pre-processing, we employed the Hampel filter to eliminate artifacts and smooth the data, a technique prevalent in pupillary research. We also accounted for the pupil light reflex, maintaining consistent environmental luminance during the experiment, ensuring that only display luminance affected the pupillary response. However, since display luminance varied between the monitor and headset lens, we developed an algorithm to compute the luminance received by the eyes based on the RGB values of all pixels. Luminance was determined using the following formula [59]:

$$Luminance = \sqrt{0.299R^2 + 0.587G^2 + 0.114B^2} \qquad …Eq\ (5)$$

Subsequently, we employed a pupil diameter light response formula [60,61] to isolate pupillary changes caused solely by cognitive load variances from our experiment. We utilized the average value from the initial 90 samples as the pupil size baseline, adjusted with a subtractive baseline correction. To address individual variability in pupil light response latency, we applied the symbolic approximation (SAX) method [62]. With our method, we could monitor the real-time cognitive load changes. **Fig. 12** shows the aggregated visualization of cognitive load changes of all participants in the 3D experience space.

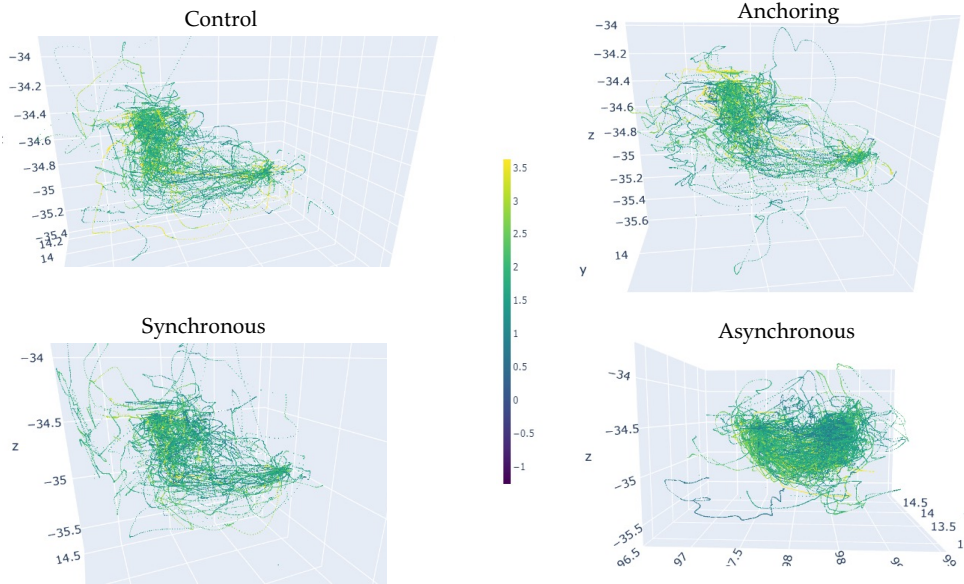

**Fig.12** Cognitive load changes in the 3D experiment space (N=41).

Ultimately, our focus centered on evaluating pupil dilation magnitude, given its documented association with cognitive load escalation. Specially, we used aggregated pupil dilation (mm) as the indicator of the cognitive load. it is the total value of pupil dilation above the personal baseline over time. Aggregated pupil dilation represents total cognitive demand during a task [63-66].

$$D = \sum_{i=1}^{n} d_i \quad \ldots Eq\ (6)$$

Where $d_i$ is the pupil dilation of a frame, $n$ is the number of frames that pupil dilated, and $D$ is the aggregated pupil dilation. Our data also shows that the proposed sensory manipulation method also presents benefits in terms of cognitive load for delays up to 1s. We did not see any difference among the four conditions when all data from each trial was aggregated in a holistic analysis. However, after dividing the data of each trail into two stages: object pickup stage (20s) and object drop-off stage (20s), we found that anchoring condition led to lower cognitive load in both the object pickup stage and the object drop-off stage, see **Fig.13** and **Table 3**.

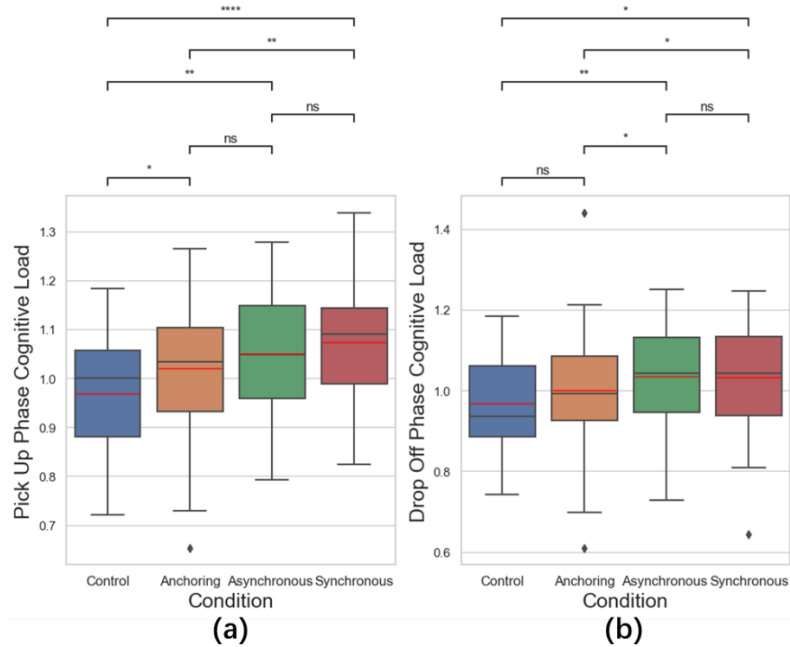

**Fig.13** Cognitive load changes in (a) object pickup and (b) drop-off stages.

| Condition Comparison | Pick Up Phase Cognitive Load | Drop Off Phase Cognitive Load |
|---|---|---|
| Control vs Anchoring | Smaller (p=0.032) | No Difference (p=0.178) |
| Control vs Asynchronous | Smaller (p=0.003) | Smaller (p=0.006) |
| Control vs Synchronous | Smaller (p<0.001) | Smaller (p=0.012) |
| Anchoring vs Asynchronous | No Difference (p=0.086) | Smaller (p=0.048) |
| Anchoring vs Synchronous | Smaller (p=0.004) | Smaller (p=0.045) |
| Asynchronous vs Synchronous | No Difference (p=0.276) | No Difference (p=0.983) |

**Table 3** Statistical results of the cognitive load metrics

Interestingly, the NASA TLX analysis did show a similar benefit of anchoring condition in terms of mental load. But the anchoring condition led to a higher level of confidence and a lower level of frustration in comparison with the synchronous condition and the asynchronous condition, as shown in **Fig.14** and **Table 4**.

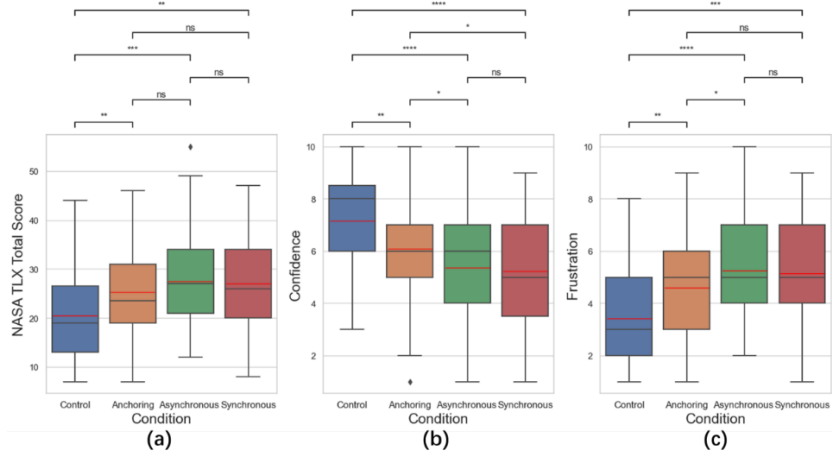

**Fig.14** SA TLX result related to (a) total score (when calculating the total score, $10 - Confidence\ score$ is used as the calculation parameter), (b) self-confidence level, and (c) frustration level for delays up to 1s. Other NASA TLX results are not shown because of the insignificant difference among the conditions.

| Condition Comparison | Total Score | Confidence | Frustration |
|---|---|---|---|
| Control vs Anchoring | Smaller (p=0.021) | Larger (p=0.007) | Smaller (p=0.004) |
| Control vs Asynchronous | Smaller (p=0.006) | Larger (p<0.001) | Smaller (p=0<0.001) |
| Control vs Synchronous | Smaller (p=0.024) | Larger (p<0.001) | Smaller (p<0.001) |
| Anchoring vs Asynchronous | No Difference (p=0.470) | Larger (p=0.024) | Smaller (p=0.033) |
| Anchoring vs Synchronous | No Difference (p=0.843) | Larger (p=0.019) | No Difference (p=0.110) |
| Asynchronous vs Synchronous | No Difference (p=0.632) | No Difference (p=0.829) | No Difference (p=0.694) |

**Table 4** Statistical results of the questionnaire metrics

## Discussion and Conclusions

Robot teleoperation, the technique of remotely controlling robots, offers the possibility of human-machine interactions in inaccessible or hazardous environments. A significant challenge within this realm is the delay between a command's issuance and its execution, particularly in long-distance operations. This delay adversely affects the operator's performance, situational awareness, and cognitive load. In the face of this challenge, the primary objective of this study was to explore a novel approach termed "induced human adaptation". Rooted in motor learning and rehabilitation principles, this method posits that strategically modifying sensory stimuli could alleviate the subjective experience of these delays and foster rapid human adaptation to them.

The experiment confirmed a variety of benefits of the proposed sensory manipulation method in teleoperation tasks with delays up to 1s. It generally confirmed that providing haptic cues coupled with the initiated action could significantly reduce time on task, no matter how much visual delay presented. It was also found that participating subjects tended to perceive a smaller visual delay when real-time haptic cues were provided. There are also benefits related to reduced cognitive load, improved perception about self-confidence and frustration levels, and more desired neural functional performance. The findings suggest that the anchoring method, i.e., providing real-time haptic feedback, has multiple performance and human functional benefits. **Table 5** summarizes the main observations.

| **Measurement** | **Metrics** | **Main observations** |
|---|---|---|
| Performance | Time on task | Anchoring (adding real-time haptic cues) could significantly reduce time on task. |
| | Positioning accuracy | No difference observed; could be due to the designed difficulty of the task. |
| Perception | Delay perception errors | Anchoring and synchronous, i.e., adding reliable haptic cues, could significantly reduce perceived visual delays. With reliable haptic cues, more than 18% of subjects reported perceived delays lower than the actual values. |
| Cognitive Load | NASA TLX | Anchoring and synchronous, i.e., adding reliable haptic cues, could reduce frustration, and increase perceived performance. |
| | Eye tracking | No difference when aggregated; but there are differences at different time points (higher load when drop-off) and for different subtasks (higher load for longer distance); suggesting a dynamic cognitive load measure is better than retrospective measures. |

**Table 5.** Main findings of the experiment (for delays up to 1s)

In addressing the issues of time delays in robot teleoperation, this paper introduces an innovative alternative to the conventional automation design and training paradigm as induced human adaptation. Drawing inspiration from motor learning and rehabilitation, the study confirmed that modified sensory stimulation, synchronously paired with motor actions, can effectively diminish the subjective sensation of time delays. This paves the way for swift human adaptation to time-delayed teleoperation, setting the requisites for extensive training or intricately designed automation/robotic systems. Through this novel approach, the study not only confronts the challenges posed by teleoperation delays but also adds a valuable dimension to the existing body of knowledge in the realm of robotics and automation. The findings help mitigate the risk of inadequate design of human and robotic interaction per NASA. Specifically, it provides alternative system designs for effective teleoperation that address variable transmission latencies via the expedited human adaptation. A recent completed NASA research [67] established quantitative models of the interaction between latency and task difficulty. It provided reference data in which latency may be traded off against rotational or other types of difficulty [67]. The data serves as a potential mitigation to time delays based on predicted performance outcome. This study further provides new understanding about how to expedite operator's adaptation to various time delays when compensation mechanisms are absent, and how subject perception of time delays can be proactively manipulated with existing systems. This study is expected to fill the gap on human adaptation knowledge.

## Acknowledgements
This material is supported by the National Science Foundation (NSF) under grant 2024784 and The National Aeronautics and Space Administration (NASA) under grant 80NSSC21K0845. Any opinions, findings, conclusions, or recommendations expressed in this article are those of the authors and do not reflect the views of the NSF or NASA.

## DATA AVAILABILITY
All data used in this paper can be found at: https://www.dropbox.com/sh/o5iug84vlnixd3u/AAC6oQOrfRx-QrxaRToJ-gWPa?dl=0

## Author Contributions Statement
Jing Du and Tianyu Zhou wrote the main manuscript text and William Vann performed the experiment. Tianyu Zhou, Yang Ye and Qi Zhu designed the system and performed data analysis. All authors reviewed the manuscript.